\pdfoutput=1

\documentclass[11pt]{article}

\usepackage[final]{acl}

\usepackage{times}
\usepackage{latexsym}

\usepackage[T1]{fontenc}

\usepackage[utf8]{inputenc}

\usepackage{microtype}

\usepackage{inconsolata}

\usepackage{graphicx}

\usepackage{hyperref}       
\usepackage{booktabs}       
\usepackage{amsfonts}       
\usepackage{nicefrac}       
\usepackage{microtype}      
\usepackage{xcolor}         
\usepackage{adjustbox}
\usepackage{xspace}
\usepackage{amsmath}
\usepackage{mathtools}
\usepackage{amssymb}
\usepackage{graphicx}

\title{HelpSteer3: Human-Annotated Feedback and Edit Data to Empower Inference-Time Scaling in Open-Ended General-Domain Tasks}

\newcommand{\huggingface}{\raisebox{-1.5pt}{\includegraphics[height=1em]{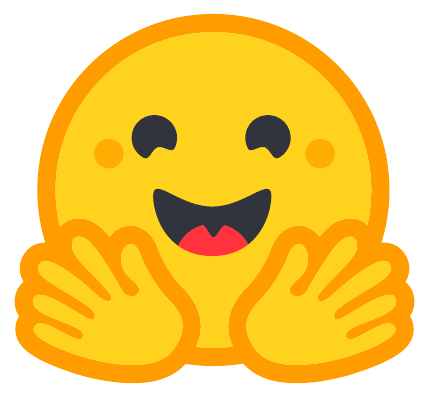}}}

\author{Zhilin Wang,  Jiaqi Zeng, Olivier Delalleau, Daniel Egert, \\
\textbf{Ellie Evans, Hoo-Chang Shin, Felipe Soares, Yi Dong, Oleksii Kuchaiev} \\
  NVIDIA \\
  \texttt{\{zhilinw, jiaqiz\}@nvidia.com} \\
  \huggingface{} \textbf{Data (CC-BY-4.0): }\url{https://huggingface.co/datasets/nvidia/HelpSteer3} \\
  \huggingface{} \textbf{Models:} 
  \href{https://huggingface.co/collections/nvidia/llama-nemotron-feedback-edit-inference-time-scaling-67d921d861f369297a46bb08}{hf.co/collections/nvidia/llama-nemotron-feedback-edit-inference-time-scaling} \\
  }

\begin{document}
\maketitle

\begin{abstract}
Inference-Time Scaling has been critical to the success of recent models such as OpenAI o1 and DeepSeek R1. However, many techniques used to train models for inference-time scaling require tasks to have answers that can be verified, limiting their application to domains such as math, coding and logical reasoning. We take inspiration from how humans make first attempts, ask for detailed feedback from others and make improvements based on such feedback across a wide spectrum of open-ended endeavors. To this end, we collect HelpSteer3 data to train dedicated Feedback and Edit Models that are capable of performing inference-time scaling for open-ended general-domain tasks. In our setup, one model generates an initial response, which are given feedback by a second model, that are then used by a third model to edit the response. We show that performance on Arena Hard, a benchmark strongly predictive of Chatbot Arena Elo can be boosted by scaling the number of initial response drafts, effective feedback and edited responses. When scaled optimally, our setup based on 70B models from the Llama 3 family can reach SoTA performance on Arena Hard at 92.7 as of 5 Mar 2025, surpassing OpenAI o1-preview-2024-09-12 with 90.4 and DeepSeek R1 with 92.3.

\end{abstract}

\section{Introduction}

\paragraph{Humans learn and improve through rich feedback from others} Take the example of writing an academic paper. Authors typically write a first draft, seek feedback from their colleagues (informally or through peer review) and improve their paper based on the feedback. Similarly, when writing new code, developers commonly create a Pull/Merge Request, seek feedback from other developers and improve their code based on the feedback. Outside of writing a paper/code, even as we are taking significant life decisions (e.g. choosing a job or PhD program), we first develop an initial inclination, seek feedback from family/friends and update our decision in light of their feedback. In each of these cases, we benefit from \textit{rich} feedback that goes beyond whether the initial draft is good/bad, and discusses why specific areas might be of concern and how those can be improved.

\begin{figure}[t]
    \centering
    \includegraphics[width=7.6cm]{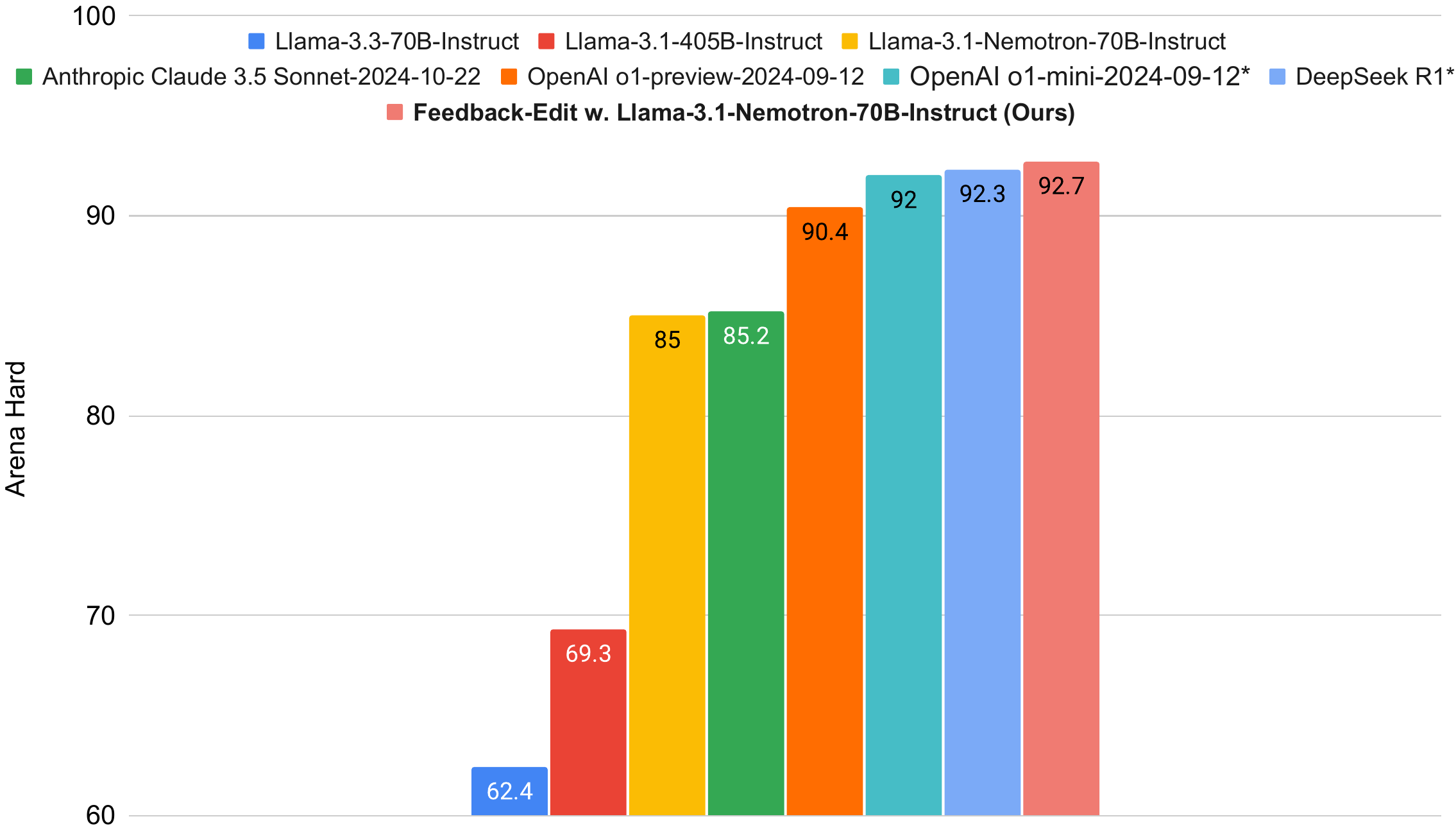}
    \caption{Dedicated Feedback and Edit Models trained with HelpSteer3 enable Llama-3.1-Nemotron-70B-Instruct to reach SoTA performance on Arena Hard at 92.7 as of 5 Mar 2025, compared to 90.4 achieved by OpenAI o1-preview-2024-09-12 \citep{arenahardauto} and 92.3 by DeepSeek R1 \citep{deepseekai2025deepseekr1incentivizingreasoningcapability} * means model specialized in Math, Coding and Logic problems, which are highly represented in Arena Hard.}
    \label{fig:front_fig}
\end{figure}

\paragraph{Reinforcement Learning from Human Feedback is not enough} While feedback from humans has inspired methods to align large language models such as Reinforcement Learning from Human Feedback (RLHF), we have thus far only utilized limited forms of such feedback. Early works \citep{ouyang2022training, bai2022training} use information on whether a response is preferred in terms of its helpfulness over the other. Subsequent works expand upon the single dimension of preferences by adding dimensions such as factual correctness, creativity, complexity and coherence \citep{wu2023finegrainedhumanfeedbackgives, kopf2023openassistant, cui2023ultrafeedback, wang-etal-2024-helpsteer}. 

However, such fixed dimensions might still be insufficient to capture the holistic feedback for a response. For instance, if a response is marked down on factual correctness, it might not be specific enough to infer where such mistakes were made, especially if the response contains thousands of words. 
\citet{welleck2022generatingsequenceslearningselfcorrect} supports this belief by finding that oracle natural language feedback (i.e. based on differences from the ground truth correct answer) can be more helpful than a rating-only scalar feedback on math, constrained natural language generation and toxicity reduction tasks but it remains unclear how useful natural language feedback will be in tasks without ground truth correct answers, which are common among general domain tasks.

\paragraph{Inference-Time Scaling by training models to think before responding is insufficient}

A popular technique for inference-time scaling is to train models to think before responding. This has been used by many models including QwQ-32B-Preview \citep{qwq}, Gemini 2.0 Flash Thinking \citep{geminiflashthinking} and Deepseek R1 \citep{deepseekai2025deepseekr1incentivizingreasoningcapability}. A reproduction attempt \citep{qin2024o1replicationjourneystrategic} of OpenAI o1 \citep{openaio1} uses a similar approach. While this approach is promising, it should not be regarded as the only axis for inference-time scaling. Techniques utilized to induce such behavior typically require that tasks have answers that can be verified, restricting their application to domains such as math, multiple-choice question answering and competitive coding situations. These techniques include Reinforcement Learning with verifiers \citep{deepseekai2025deepseekr1incentivizingreasoningcapability, lambert2024tulu3pushingfrontiers}, Process Reward Models \citep{lightman2023letsverifystepstep}, Generative Verifiers \citep{zhang2024generativeverifiersrewardmodeling, zhang2024smalllanguagemodelsneed}, Monte-Carlo Tree Search \citep{zhang2024rest}, Self-Consistency \citep{wang2023selfconsistencyimproveschainthought} and Execution Feedback \citep{gehring2024rlefgroundingcodellms}. 

However, there exists a large set of open-ended but challenging problems for which no such ground-truth verifiable answers exists. Such problems include proposing good research ideas \citep{si2024llmsgeneratenovelresearch}, writing a speech for important events or even coming up with an effective approach for delivering a complex software product. Interestingly, \citet{openaio1} found that GPT-4o (without thinking before responding) is preferred over OpenAI o1 in the domain of personal writing, suggesting that thinking-before-responding is not suited for some tasks. Therefore, there needs to be complementary approaches for inference-time scaling.

\paragraph{Efficacy of LLMs in providing feedback and making edits to their own responses is unclear} Given the value of feedback and refinement loops for inference-time scaling, the most straightforward solution is to simply prompt an instruction-following LLM to provide feedback and/or editing on its own \citep{buildingeffectiveagents, yuan2024selfrewardinglanguagemodels, wu2024metarewardinglanguagemodelsselfimproving}. While this is possible, LLMs that were not specifically trained might not be able to provide effective feedback leading to minimal improvements relative to high quality feedback \citep{chen2023teachinglargelanguagemodels, tyen2024llmsreasoningerrorscorrect}. A supporting observation is that using feedback from general instruction-following models \citep{ye2024improvingrewardmodelssynthetic, ankner2024critiqueoutloudrewardmodels} is less useful for preference modeling compared to using feedback from a model that has been trained (using distant supervision signals) to generate better feedback \citep{yu2024selfgeneratedcritiquesboostreward}. 

There are some reports that LLMs can self-correct when given high quality feedback for code/reasoning-related \citep{chen2023teachinglargelanguagemodels, shi2024languagemodelssolveolympiad, tyen2024llmsreasoningerrorscorrect, chen2024improvingcodegenerationtraining, lin2024criticbenchbenchmarkingllmscritiquecorrect} and documented-grounded/summarization tasks \citep{saunders2022selfcritiquingmodelsassistinghuman, wadhwa2024learningrefinefinegrainednatural}. However, we are not aware of any study on whether LLMs can productively edit their responses based on provided feedback in general domain tasks. Relatedly, \citet{ji2024alignerefficientalignmentlearning} show that a specifically trained model can improve model responses to general domain tasks without using feedback, but its efficacy relative to improvements conditioned on feedback is under-explored. 

\paragraph{Our Solution: HelpSteer3 Enables Training of Dedicated Feedback-Edit Models that are Inference-Time Scalable for SOTA performance}

\begin{enumerate}
    \item We release HelpSteer3 data (CC-BY-4.0-licensed) from 7000+ annotators across 80+ regions in diverse, open-ended general-domain tasks. Annotators give textual feedback relating to response helpfulness and use such feedback to edit the original responses.
    \item We train dedicated Feedback and Edit models with HelpSteer3 data that can be used at inference time to improve model responses to open-ended general-domain tasks.
    \item We show that the Feedback-Edit system can be effectively scaled by generating more initial responses, feedback and edits. Together, an optimal setup based on 70B models from Llama 3 family can reach 92.7 on Arena Hard, surpassing OpenAI o1-preview-2024-09-12 with 90.4  and DeepSeek R1 with 92.3.
\end{enumerate}

\section{Dataset}

\subsection{Data Collection}

\paragraph{Prompt Curation}
Our data consists of prompts from several categories (General, STEM, Coding and Multilingual), similar to Llama 3 data \citep{dubey2024llama3herdmodels}. Across all tasks, we perform a filter to exclude potentially unsafe prompts/responses  (e.g. harmful content, illegal activities, profanity, bias and stereotyping) and ask annotators to skip tasks missed by our filter, in addition to tasks containing Personally Identifiable Information (e.g. name, address, SSN, email, phone numbers).

Following HelpSteer2 \citep{wang2024helpsteer2opensourcedatasettraining}, we sourced Coding and Multilingual prompts from ShareGPT \citep{sharegpt2023}, a dataset curated from user conversations with ChatGPT that are voluntarily shared.
We chose ShareGPT because HelpSteer2 has demonstrated promise as a high-quality preference modeling dataset but previously lacked Coding and Multilingual prompts. 
For General and STEM prompts, we use WildChat \cite{zhao2024wildchat1mchatgptinteraction}, a dataset similar to ShareGPT but significantly larger (approximately 1 million prompts), to minimize potential overlap with HelpSteer2. Further details on prompt language classification and preprocessing steps are provided in Appendix \ref{appendix:prompt_preprocessing}.

\paragraph{Response Generation}\label{sec:response_generation}
Following HelpSteer2 \citep{wang2024helpsteer2opensourcedatasettraining}, we generate two responses per prompt using a diverse selection of models, including Nemotron 4 340B Instruct \citep{nvidia2024nemotron4340btechnicalreport}, Mistral Large 2 \citep{mistrallarge2}, Mistral 7B-Instruct-v0.3 \citep{mistral7bv03}, Mixtral 8x22B Instruct \citep{mixtral822b}, Mixtral 8x7B Instruct \citep{jiang2024mixtral}, Mistral NeMo 12B \citep{mistralnemo}, Codestral 22B \citep{codestral}, Gemma 2 (2B, 9B, 27B) \citep{gemmateam2024gemma2improvingopen}, Gemma 2B \citep{gemmateam2024gemmaopenmodelsbased}, Phi 3 (Mini, Small, Medium) \citep{abdin2024phi3technicalreporthighly}, IBM Granite (8B and 34B) \citep{mishra2024granitecodemodelsfamily} and Snowflake Arctic \citep{snowflakearctic}.
We selected these models because they were popular at the start of our annotation exercise and have commercially permissive licenses\footnote{Codestral 22B and Mistral Large 2 were used following a contractual agreement with Mistral to use and release the responses by these models under a commercially permissive license, which we chose as CC-BY-4.0.}. 
All models were accessed through the NVIDIA API Catalog \citep{nvidiaapicatalog}. 
To enhance the compatibility of our Feedback and Edit Models, we deliberately included models with varying sizes and capabilities, prioritizing larger models for generating more responses.
Responses were generated with temperature 0, top p 0.9 and maximum tokens of 3072.

\paragraph{Multi-turn Prompt Completion}
Following HelpSteer2 \citep{wang2024helpsteer2opensourcedatasettraining}, we include multi-turn prompts because users might want to apply Feedback-Edit loop for follow-up prompts. Instead of using ChatGPT generated assistant turns from ShareGPT/WildChat, we use the models used for response generation to generate intermediate assistant turns. This was done to avoid using any ChatGPT response in our dataset curation process.

\paragraph{Feedback Annotation}\label{sec:feedback_annotation}
Annotators are asked to provide free-text feedback on the overall helpfulness of the last assistant response in a conversation. Concretely, annotators are asked to provide a feedback of 2-10 sentences (50 - 250 words) that starts with `The response is \{not / slightly / partially / mostly / perfectly\} helpful'. For Multilingual data, we specifically encourage the annotators to provide feedback if the response language is not the same as the prompt language or the language the user expects the output in (e.g. language-learning tasks). For Coding data, we encourage annotators to provide feedback on whether there are sufficient comments to describe the code and whether code adheres to best practices following \citet{googlestyleguide}.

All feedback is provided in English, including for Multilingual responses. Annotators evaluate two responses from different models sequentially, with the ability to review both before finalizing their feedback. This process helps calibrate their assessments and prevents overly harsh feedback for minor issues.
Following HelpSteer2 \citep{wang2024helpsteer2opensourcedatasettraining}, we assign 3–5 annotators per task to reduce subjectivity and ensure more balanced feedback across model responses. 
We collaborate with a data annotation vendor, Scale AI, to recruit annotators for General, STEM, and Coding tasks. 
STEM and Coding tasks have stricter inclusion criteria, requiring annotators to have relevant qualifications, such as undergraduate degrees in related fields or software engineering experience. 
These tasks, along with their associated prompts and responses, are conducted in English only.
For Multilingual tasks, we partner with a separate vendor, Translated.
Across both vendors, we recruit over 7,000 annotators from more than 80 countries, with details provided in Table \ref{tab:geographic}.
Annotators are encouraged to use internet resources for fact-checking when necessary but prohibited from using LLMs. 

\paragraph{Response Editing}\label{sec:response_editting}

We consolidate feedback for each task and send them to separate pools of annotators to edit responses based on the provided feedback. Our qualitative inspection of early feedback suggests that outlier annotations (i.e. an annotator who finds a response to be slightly helpful when four annotators find it perfectly helpful) tend to be mistaken. Therefore, we follow HelpSteer2 to only use feedback from three annotators that agree most with one another, in terms of their overall helpfulness judgment.
We note that feedback from three annotators is preferable to feedback from a single annotator, as different people may identify distinct areas for improvement, even if their overall assessments of helpfulness are similar.
Additionally, we find that responses rated as either perfectly helpful or not helpful by most annotators are not suitable for editing.
The former cannot be meaningfully improved while the latter needs to be rewritten entirely. Therefore, we exclude such responses. 

\subsection{Data Pre-processing}\label{sec:data_preprocessing}

Using the collected data, we formulate three datasets for model training:

\paragraph{Feedback Demonstration}\label{sec:feedback_imitation}
aims to show models how humans provide feedback in order to train them to imitate such feedback, given a user prompt and a model response. With the data collected for \textit{Feedback Annotation}, we perform a few data cleaning steps. First, we retain only the three feedback per prompt that agree most to filter out outliers.
An advantage of using multiple feedback per response is that it increases the diversity of potential model responses when doing sampling, instead of over-fitting to a single feedback. Then, we remove prompts for which the three most agreeing prompts have too large a disagreement (e.g. one finds the response to be perfectly helpful while another finds it to be slightly helpful). Such a large degree of disagreement typically suggests that one or more of the annotators overlooked or misunderstood an important aspect of the response. 

\paragraph{Edit Demonstration} seeks to show how humans edit responses by training the model to generate an edited response given a user prompt, a model response and a set of feedback. We first filter out any feedback that starts with that `The model response is perfectly helpful' because such type of feedback does not contain any suggestion for improvement. We also want to exclude feedback that was not used in the editing process to ensure the model learns to apply every piece of feedback. However, we did not collect any data that explicitly specifies which feedback were (actually) used to edit the response. Therefore, we use a distant supervision signal to identify this instead. Specifically, we use the change summary that each edit annotator needed to write to accompany their edits. We then prompt Llama-3.3-70B-Instruct \citep{llama33} to see if the change summary addresses the issues mentioned in the feedback (prompt template in Appendix \ref{appendix:prompt_templates}). Feedback which were not mentioned in the change summary were excluded. Finally, we include all permutations of linearized feedback from the set of feedback, in order to teach the model the order-independence of the feedback when performing the edit.

\paragraph{Edit Preference}\label{sec:edit_preference} is used to train a model to distinguish good edits from unsatisfactory edits. The first type of unsatisfactory edits are bad edits that are not based on the feedback, but instead on what an edit annotator thought was needed (see Appendix \ref{appendix:bad_edit_reasoning} for details). This was a common issue during our initial annotation. We identified tasks that have a bad edit as well as a good edit (i.e. the task was reworked later on such that it follows the provided feedback). The second type of undesirable edits is when no edit was made (i.e. the edited response simply copies the original model response). While this was not seen in the human annotated data, we observed this phenomenon in a substantial fraction of a model trained on the Edit Demonstration dataset. We believe this might be related to how Supervised Fine-tuning calculates loss at a token level and hence the model might learn a short-cut that simply copying the original response can minimize the loss for most tokens since the human edits preserve most initial tokens. We create an Edit Preference dataset by having these two types of unsatisfactory edits in a 1:1 ratio, paired with the corresponding Good Edit. Only data relating to General/STEM subsets is available for this task.

\subsection{Data Analysis}\label{sec:preference_analysis}

\begin{table}[ht!]
\centering
\begin{adjustbox}{max width=\columnwidth, scale=1
}
\begin{tabular}{lccccc}
\toprule

& \textbf{Overall} & \textbf{General} & \textbf{STEM} & \textbf{Coding} & \textbf{Multilingual} \\ 

\midrule
\textbf{Feedback Demonstration} \\
\midrule

No. of Responses (\%) & 81642 (100)  & 37276 (45.7)  & 9836 (12.0)  & 18404 (22.5) & 16126 (19.8)\\
Feedback Len. in Chars. (std.)  & 437.8 (163.5) & 438.3 (147.6) & 446.3 (151.9) & 518.6 (190.8) & 339.2 (110.4)\\
\midrule
\textbf{Edit Demonstration} \\
\midrule
No. of Original Responses (\%) &  14461 (100) & 4848 (33.5) & 1196 (8.3) & 3548 (24.5) & 4869 (33.7) \\
No. of Feedback per Edit (std.)   &   2.3 (0.9) & 2.4 (0.8) & 2.3 (0.8) & 2.5 (0.8) & 2.0 (0.9) \\
Original Response Len in Chars. (std.) &  1642.7 (1228.6) & 1714.6 (1248.2) & 1609.2 (1098.0) & 2243.7 (1202.8) & 1141.4 (1030.1) \\
Edited Response Len in Chars. (std.)  &  1813.1 (1570.9) & 1748.9 (1399.7) & 1581.6 (1195.3) & 3025.7 (1785.4) & 1050.4 (1017.1) \\
> $\Delta$ in Response Len in Chars. (std.) & 170.4 (899.5) & 34.3 (758.3) & -27.6 (693.7) & 782.0 (1191.1) & -90.9 (556.5) \\

\bottomrule
\end{tabular}
\end{adjustbox}
\caption{Descriptive Statistics for Feedback and Edit Demonstration datasets.}
\label{tab:descriptive_feedback_imitation}
\end{table}

Table \ref{tab:descriptive_feedback_imitation} shows that more than half of the data from both Feedback and Edit Demonstration datasets are from specialist annotators (STEM, Coding and Multilingual). This means that the datasets contains diverse feedback and edits, which can help train models to be useful across domains. The dataset also contains 14 programming and 13 natural languages with detailed distributions in Appendix \ref{appendix:further_dataset_statistics}. An example of General task is in Appendix \ref{appendix:example}.

\paragraph{Feedback Demonstration} Multilingual feedback (339.2 characters) tends to be shorter while coding feedback (518.6 characters) is longer than overall (437.8 characters). Our qualitative inspections of the multilingual feedback suggests that multilingual prompts tend to be straightforward and do not typically require annotators to perform extensive fact checking in order to provide feedback. Therefore, feedback annotators might have less content to write. On the other hand, coding prompts often require annotators to check the code implementation (often using real-world software libraries). This means that feedback annotators can point out concrete issues with the code, giving them more to write in their feedback. 

\paragraph{Edit Demonstration} on average contains slightly fewer than 3 feedback per response (2.3), as a small fraction has been removed due to the feedback mentioning that the response is perfect (and hence no need to edit) or the change summary accompanying the edit not mentioning issues raised by the feedback (and likely ignored by the annotator). Original responses are moderately long at 1642.7 characters and similar in length to the responses in HelpSteer2 with 1492.6 characters \citep{wang2024helpsteer2opensourcedatasettraining}. Coding responses tend to be substantially longer at 2243.7 characters, likely due to extensive white-spaces in languages such as Python. Conversely, multilingual responses have fewer characters (1141.4) possibly because many languages in the dataset (see Table \ref{tab:descriptive_feedback_languages}) use scripts different from the Roman alphabets used in English.

\begin{table}[ht!]
\centering
\begin{adjustbox}{max width=\columnwidth, scale=1
}
\begin{tabular}{lccc}
\toprule

& \multicolumn{1}{c|}{\textbf{Overall}} & \multicolumn{1}{c|}{\textbf{General}} & \multicolumn{1}{c}{\textbf{STEM}}\\
\midrule
No. of Original Responses (\%) & 3274 (100)  & 2605 (79.6) & 669 (20.4) \\
No. of Feedback per Edit (std.) & 2.3 (0.8)  & 2.3 (0.8) & 2.3 (0.9) \\
Original Response Len in Chars. (std.) & 1716.8 (1236.4) & 1746.7 (1274.0) & 1600.1 (1069.5) \\
Bad Edited Response Len in Chars. (std.) & 1594.6 (1322.8)  & 1649.0 (1390.8) & 1383.0 (987.7)\\
> $\Delta$ in Bad Edit Len in Chars. (std.) &  -122.1 (804.1)  & -97.8 (822.3) & -217.0 (720.8) \\ 
Good Edited Response Len in Chars. (std.)  &  1749.3 (1377.6)  & 1788.9 (1428.8) & 1595.2 (1144.0) \\
> $\Delta$ in Good Edit Len in Chars. (std.) & 32.5 (727.0)  & 42.1 (741.1) & -4.9 (667.8) \\

\bottomrule
\end{tabular}
\end{adjustbox}
\caption{Descriptive statistics for Edit Preference data.}
\label{tab:descriptive_feedback_preference}
\end{table}
Within the General, STEM and Multilingual subsets, the edited responses have similar length as the original responses (within 100 characters). On the Coding subset however, there is a substantial increase in average length by 782 characters. This is likely due to the Edit annotator adding more comments and explanation to the code or including handling of edge cases not included in the original response (e.g. making a solution work on Windows in addition to Unix systems).

\paragraph{Edit Preference} In Table \ref{tab:descriptive_feedback_preference}, bad edits tend to lead to a decrease in response length by an average of 122.1 characters.
This is likely because such edits are typically based on the edit annotator's own opinion of what should be changed and the averse incentive for annotators to shorten responses since it's relatively easier to delete content than to write new content. On the other hand, good edits result in a minimal change of response length (an increase of 32.5 characters, with std. of 727), suggesting that annotators of good edits were willing to either include missing content or exclude superfluous writing, depending on provided feedback.

\section{Experiments}

\subsection{Evaluation}\label{sec:evaluation}

\paragraph{Metrics} Following \citet{wang2024helpsteer2preferencecomplementingratingspreferences, dong2024rlhf, meng2024simpo}, we measure the helpfulness of models to general-domain prompts using three popular metrics: AlpacaEval 2.0 Length Controlled \citep{dubois2024length}, GPT-4-Turbo MT Bench \citep{zheng2023judging} and Arena Hard \citep{arenahard2024}. AlpacaEval contains Easy prompts, MT Bench contains Medium-difficulty prompts (including multi-turn) and Arena Hard contains the hardest prompts. All responses are generated greedily. To give a sense of average response length, we report the mean number of characters in MT Bench responses. MT Bench is also used as a validation metric for checkpoint selection.
Further details are in Appendix \ref{appendix:aligned_model_evaluation}.

\paragraph{Feedback and Edit Generation} We generate 10 feedback per initial response with a temperature of 0.7 and top p of 0.9. We then exclude any feedback which finds the initial response `perfectly helpful', since such feedback does not contain information relating to improvement(s) for the Edit model to make. Out of the remaining feedback, we randomly choose up to 3 feedback for generating an edit of the initial response. The choice of generating 10 feedback was made to balance the amount of feedback required against having most responses having 3 feedback after filtering. We then greedily generate the edited response conditioned on the prompt, initial response, and feedback. Prompt templates for feedback and edit generation can be found in Appendix \ref{appendix:prompt_templates}.

\subsection{Model Training}\label{sec:model_training}

We use the Llama 3.3 70B Instruct model \citep{llama33} to initialize Feedback and Edit Supervised Fine-Tuning (SFT)  model training. Edit Reward Model (RM) and Reinforcement Learning (RL) model are initialized from the Edit SFT Model. 
Each dataset is split into 95\% train and 5\% validation.
Prompt templates are available in Appendix \ref{appendix:prompt_templates} and further details are in Appendix \ref{appendix:training_details}.

\paragraph{Feedback SFT} We perform SFT on the Feedback Demonstration dataset. We train for 1 epoch (with 1,858 steps of global batch size of 128) and save checkpoints every 200 steps. The checkpoint with
best validation performance is at step 1,400. 

\paragraph{Edit SFT} We perform SFT on the Edit Demonstration dataset. We train on 1 epoch (with 385 steps of a global batch size of 128) and save a checkpoint every 100 steps. The checkpoint with best validation performance is at step 385.

\paragraph{Edit RM} We perform Bradley-Terry modeling \citep{ouyang2022training, bai2022training} on the Edit Preference dataset. We train up to 1 epoch (94 steps of global batch size 128) and save a checkpoint every 10 steps. The checkpoint with the lowest validation loss is at step 80. While the dataset is leaning small, we kept to 1 epoch because Bradley-Terry (BT) style Reward Models over-fit when trained beyond \citep{wang2024helpsteer2preferencecomplementingratingspreferences, zhu2024iterativedatasmoothingmitigating}.
One design we implemented was that each step only contains 32 distinct `prompt-initial-response-feedback' tuples with one pair of (bad edit, good edit) and another pair of (no edit, good edit). Compared to having (bad edit, good edit) and (no edit, good edit) pairs in separate global batches, this setting led to slightly lower validation loss and higher validation accuracy. We believe this is advantageous to support the RM to jointly optimize for good edits against bad and no edits in one backward pass, while preventing multiple gradient updates from having identical 'good edit' samples in separate global batches, which risks over-fitting BT RMs.

\paragraph{Edit RL} Following \citet{wang2024helpsteer2preferencecomplementingratingspreferences}, we perform REINFORCE Leave One Out \citep{ahmadian2024basicsrevisitingreinforcestyle} on the Edit SFT model guided by the Edit RM. We train up to 1 epoch (47 steps of a global/rollout batch size of 64) and save a checkpoint every 5 steps. We select the checkpoint at step 45 which shows best validation performance.

\section{Results}

\begin{table}[ht!]
\centering
\begin{adjustbox}{max width=\columnwidth, scale=1
}
\begin{tabular}{lcccc}
\toprule

& MT Bench  & Mean Response  & AlpacaEval  & Arena Hard \\
& (GPT-4-Turbo)  & Length (Chars.) & 2.0 LC (SE) & (95\% CI) \\

\midrule
\textbf{Llama-3.1-Nemotron-70B-Instruct} & 8.98 & 2199.8 & 57.6 (1.65) & 85.0 (-1.5, 1.5) \\
\midrule
+ Feedback + Edit & \textbf{9.16} & 2614.4 & \textbf{62.8} (1.30) & \textbf{87.0} (-1.5, 1.7) \\

\midrule
\textbf{Llama-3.3-70B-Instruct} & 8.29 & 1827.6 & 35.0 (1.45) & 62.4 (-2.5, 2.5) \\
\midrule
+ Feedback + Edit & 9.07& 2362.1& 36.9 (1.50) & 74.8 (-1.7, 2.4)\\
\midrule
\textbf{External Baselines}  \\
\midrule
Llama-3.1-70B-Instruct & 8.22 & 1728.6 & 38.1 (0.90) & 55.7 (-2.9, 2.7) \\
Llama-3.1-405B-Instruct &  8.49 & 1664.7 & 39.3 (1.43) & 69.3 (-2.4, 2.2) \\ 
Claude-3-5-Sonnet-20240620 & 8.81 & 1619.9  & 52.4 (1.47) & 79.2 (-1.9, 1.7)\\
GPT-4o-2024-05-13 & 8.74 & 1752.2 & 57.5 (1.47) & 79.3 (-2.1, 2.0) \\

\bottomrule
\end{tabular}
\end{adjustbox}
\caption{Applying Feedback and Edit models with various Instruct models. Higher is better for all metrics, except Length. External Baselines numbers are taken from \citet{wang2024helpsteer2preferencecomplementingratingspreferences}. Additional comparisons are available in Appendix \ref{appendix:additional_baselines}.
}
\label{tab:auto_eval}
\end{table}

As shown in Table \ref{tab:auto_eval}, using Feedback + Edit to perform inference-time scaling on both Llama-3.1-Nemotron-70B-Instruct \citep{nemotron70binstruct} and Llama-3.3-70B-Instruct model substantially improves performance across MT Bench, AlpacaEval and Arena Hard. In addition to Llama-3.3-70B-Instruct, we chose to use Llama-3.1-Nemotron-70B-Instruct because it is the strongest open-source 70B model based on Arena Hard, AlpacaEval 2.0 LC and MT Bench \citep{wang2024helpsteer2preferencecomplementingratingspreferences}. While there is a slight increase in mean response length, both models have an increased AlpacaEval 2 Length Controlled score suggesting that the model responses are improved, even when controlled for length.

\section{Ablation Studies}

\begin{table}[ht!]
\centering
\begin{adjustbox}{max width=\columnwidth, scale=1
}
\begin{tabular}{lcccc}
\toprule
& MT Bench  & Mean Response  & AlpacaEval  & Arena Hard \\
& (GPT-4-Turbo)  & Length (Chars.) & 2.0 LC (SE) & (95\% CI) \\
\midrule
\textbf{Llama-3.1-Nemotron-70B-Instruct} & 8.98 & 2199.8 & 57.6 (1.65) & 85.0 (-1.5, 1.5) \\
\midrule
+ Feedback + Edit & 9.16 & 2614.4 & 62.8 (1.30) & 87.0 (-1.5, 1.7) \\

\midrule 
\multicolumn{5}{l}{\textbf{Self-Feedback and Self-Edit}} \\
\midrule
+ Self-Feedback + Self-Edit & 9.11 & 3587.0 &  64.6 (1.22) & 84.6 (-1.5, 1.2) \\
+ Feedback + Self-Edit & 8.94 & 2320.6 & 66.2 (1.40) & 85.4 (-1.6, 1.4)\\

\midrule
\multicolumn{5}{l}{\textbf{Removing RL from Edit Training}} \\
\midrule
 + Feedback + Edit w/o RL & 9.12 & 2284.4 & 64.4 (1.35) & 86.4 (-1.5, 1.5) \\
 
\midrule
\multicolumn{5}{l}{\textbf{Importance of Feedback for Edit}} \\
\midrule
+ Edit w/o Feedback & 9.14 & 2533.6 & 67.4 (1.29) & 84.5 (-1.4, 1.4)\\

\bottomrule
\end{tabular}
\end{adjustbox}
\caption{Ablation Studies on Feedback-Edit system with Llama-3.1-Nemotron-70B-Instruct. Higher is better for all metrics, except Length.}
\label{tab:ablation}
\end{table}

To better understand the contribution of each system component, we conduct a few ablation studies:

\paragraph{Self-Feedback and Self-Edit} First, we want to probe the value of having a Feedback-Edit loop over the initial model responses. To do this, we replace the dedicated Feedback and Edit models with Llama-3.1-Nemotron-70B-Instruct prompted similarly. As seen in Table \ref{tab:ablation}, the Self-Feedback and Self-Edit setup perform better on both AlpacaEval (57.6 to 64.6) and MT Bench (8.98 to 9.11) but worse on Arena Hard (85.0 to 84.6) when compared with just the initial Llama-3.1-Nemotron-70B -Instruct model.

We believe this difference is because AlpacaEval has simple prompts (e.g. `How did US states get their names?') while Arena Hard has very challenging prompts (e.g. `how does memory affect performance of aws lambda written in nodejs') with MT Bench in between in terms of difficulty. This suggests that simply prompting a general instruction-following LLM to improve responses does not work well on challenging prompts, underpinning the importance of training dedicated feedback and edit models. We also note that the Self-Feedback model tends to give much longer feedback (around 4 to 5 times as many characters) compared to the trained Feedback model, often encouraging to generate more details (see examples in Appendix \ref{appendix:llama31nemotron_feedback} and \ref{appendix:example_feedback}). As a result, Self-Feedback and Self-Edit setup give substantially longer responses (3587.0 characters compared to the initial model with 2199.8). 

\paragraph{Feedback and Self-Edit}

To better understand whether general Instruct models such as Llama-3.1-Nemotron-70B-Instruct are constrained by their ability to give effective feedback or to edit initial responses based on the given feedback, we conduct a separate ablation using the trained Feedback model and a Llama-3.1-Nemotron-70B-Instruct prompted to self-edit given the feedback. Note that using a prompted Self-Feedback model with an Edit model is not practical since the feedback given by Llama-3.1-Nemotron-70B-Instruct is substantially different from the distribution that the Edit model was trained on, since self-feedback is 3-4 times longer (see examples in Appendix \ref{appendix:llama31nemotron_feedback} and \ref{appendix:example_feedback}).

Relative to the Feedback and Edit set-up, there is an increase in AlpacaEval (62.8 to 66.2) but a substantial drop in MT Bench (9.16 to 8.94) and Arena Hard (87.0 to 85.4). This suggests that prompted Self-Edit models struggle with editing based on given feedback to medium or hard prompts, stressing the value of a dedicated Edit model. In addition, we noticed some unexpected behavior when prompting Llama-3.1-Nemotron-70B-Instruct to edit responses based on feedback. It sometimes starts with a prefix like ``Here's the edited response:'' or ``I'm glad to hear that the response was helpful''. Further, it occasionally plans what to do based on the feedback or provides multiple alternative edits (when the edited answer is short). Such behaviors might not be useful to end users who expect straightforward model responses, similar to the responses before applying the edits.

\paragraph{Removing RL from Edit Training} We introduce RL for training the Edit Model because we observe that the SFT-only Edit frequently (around 30\%) leaves the response entirely unchanged, even though clear areas for improvement are mentioned in the provided feedback. Post-RL, we see that such plain repetition no longer occur, which also leads to an increase in mean response length on MT Bench (from 2284.4 to 2614.4 characters). We see that the medium-difficulty MT Bench increases from 9.12 to 9.16 and high-difficulty Arena Hard increases from 86.4 to 87.0 while the low-difficulty AlpacaEval slightly drops from 64.4 to 62.8, which is likely an effect of the length-correction penalty due to the slightly longer length.

\paragraph{Importance of Feedback} Finally, we want to understand whether feedback is helpful for grounding edits, drawing from \citet{ji2024alignerefficientalignmentlearning}. We train an Edit SFT\footnote{RL training w/o feedback is not feasible because we don't have signal on what a `good edit' is vs. a `bad edit' since goodness is based on how relevant the edits are to the feedback.} model using exactly the same data and hyper-parameters while excluding feedback and slightly changing the prompt template accordingly (see Appendix \ref{appendix:prompt_templates}). Relative to 
the `Feedback + Edit w/o RL' setup, Edit w/o Feedback has a similar performance on MT Bench (9.12 vs. 9.14) but a higher score on easy prompts in AlpacaEval (64.4 vs. 67.4) and a lower score on challenging prompts in Arena Hard (86.4 vs. 84.5). It is critical to note that the Edit w/o Feedback setup achieves an even lower score on Arena Hard (84.5) compared to Llama-3.1-Nemotron-70B-Instruct \textit{alone} (85.0). 

Such an observation agrees with \citet{ji2024alignerefficientalignmentlearning} which found that training an Edit model without feedback to be useful on improving AlpacaEval 2 LC of GPT-4-Turbo from 55.0 to 58.3\%, although our improvement is substantially larger (57.6 to 67.4\%). Together, this suggests that feedback is not beneficial for simple tasks (e.g. straightforward open-ended question answering) similar to those in AlpacaEval but is advantageous for more challenging tasks (e.g. solving complex real-world code tasks) in Arena Hard which requires more targeted feedback for the model to make meaningful improvements.
To concretely demonstrate the utility of diverse feedback for challenging tasks, we include select generated feedback in Appendix \ref{appendix:example_feedback}.

\section{Scaling}\label{sec:scaling}

\begin{figure}[t]
    \centering
    \includegraphics[width=7.6cm]{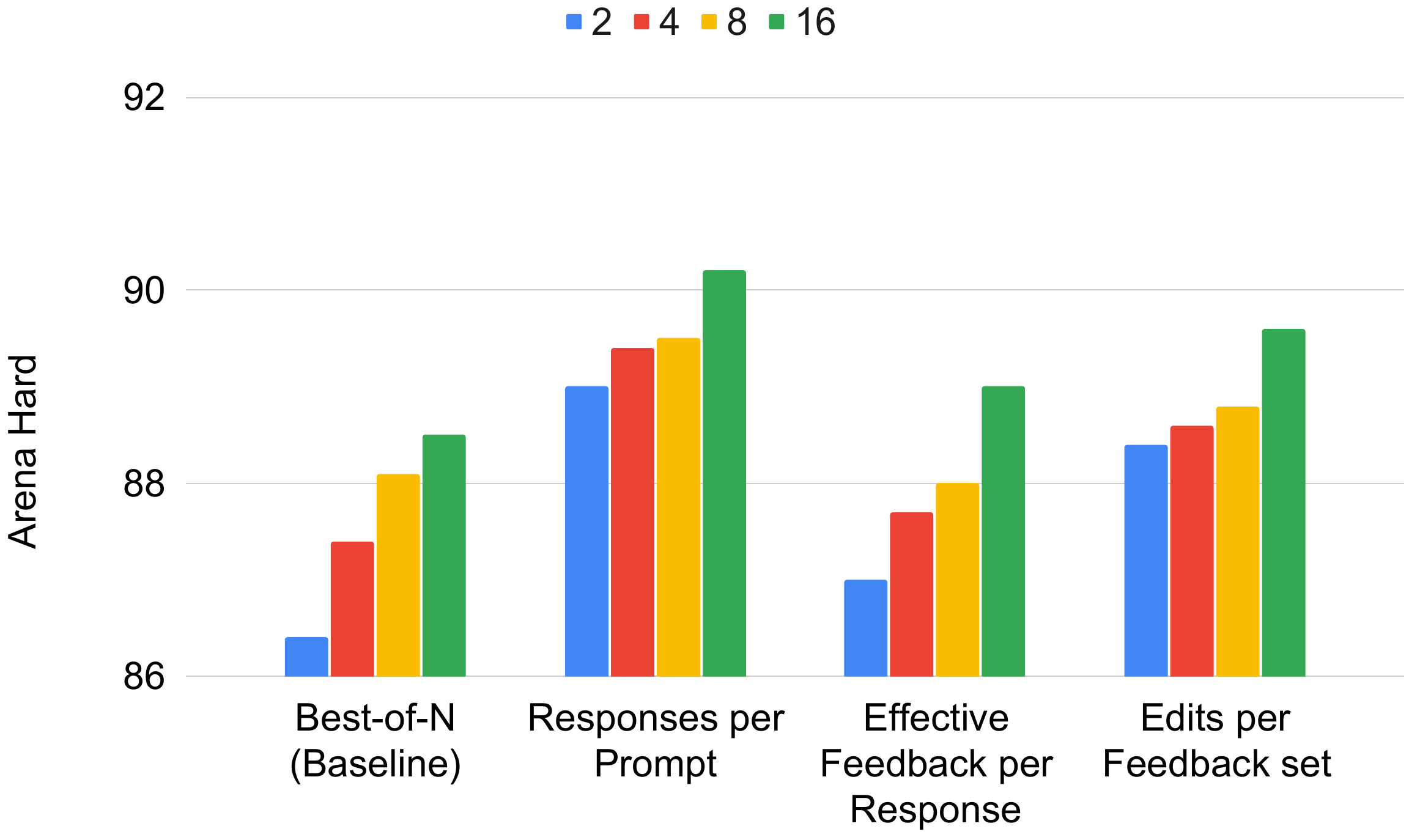}
    \caption{Feedback and Edit Models enable effective Inference-Time scaling across various dimensions.}
    \label{fig:front_page}
\end{figure}

While dedicated Feedback and Edit models can benefit model performance used at a small scale during inference-time, greater value can be unleashed by scaling up various aspects of this system. Here, we show it can be scaled at inference time by increasing the number of initial responses per prompt, effective feedback per response and edited responses per feedback set. To illustrate such capability (within compute constraint), we focus on the Arena Hard task as it is the most challenging.

\paragraph{Best-of-N (Baseline)} We perform Best-of-N Sampling, which involves temperature-sampling multiple times and using a Reward Model to pick the best generation. We generate up to 16 responses using Llama-3.1-Nemotron-70B-Instruct using temperature 0.7 and top p 0.9 and then choose the response that has the highest reward from Llama-3.1-Nemotron-70B-Reward \citep{nemotron70breward}, a top performing reward model on RewardBench \citep{rewardbenchleaderboard}. From Fig. \ref{fig:front_page}, scaling from Best of 2 to Best of 16 increases Arena Hard from 86.4 (95\% CI: -1.8, 1.7) to 88.5 (95\% CI: -1.2, 1.5).

\paragraph{Initial Responses per Prompt} We take each of these 16 responses from Best-of-N sampling and individually pass them through the Feedback-Edit system. Finally, we take the 16 edited responses and choose one with the highest reward as scored by Llama-3.1-Nemotron-70B-Reward. As shown in Fig. \ref{fig:front_page}, scaling from 2 to 16 initial responses per prompt leads to an increase in Arena Hard from 89.0 (95\% CI: -1.4, 1.2) to 90.2 (95\% CI: -1.6, 1.2).  This suggests that the Feedback-Edit setup is complementary to Best-of-N sampling.

\paragraph{Effective Feedback per Response} Our baseline setup in Sec. \ref{sec:evaluation} generates 10 feedback per response. However, when looking carefully into the feedback, we find that many feedback describe the quality of the response but do not provide constructive criticism (i.e. areas that can be improved to make the response better). This means that there are often fewer than three actionable suggestions for the Edit model to act upon. Concretely, we estimate there to be around two effective feedback. To solve this issue, we increase the number of feedback generated and re-rank feedback based on a heuristic: the number of constructive criticism  keywords\footnote{However, improve, lack, benefit, but} they contain. Therefore, we generate 16, 32 and 64 feedback respectively to obtain 4, 8 and 16 effective feedback. In addition, we find that when generating at the same temperature of 0.7, there are very few differentiated feedback beyond 16 samples. Therefore, for 32 feedback, we generate 16 with temperature 0.7 and 16 with temperature 0.6; for 64 feedback, we generate 16 feedback for each temperature - 0.5, 0.6, 0.7, 0.8\footnote{Even lower temperatures loses diversity while even higher temperature leads to frequent hallucinations}. Scaling from 2 to 16 effective feedback increases Arena Hard from 87.0 (95\% CI: -1.5, 1.7) to 89.0 (95\% CI: -1.5, 1.3).

\paragraph{Edited Responses per Feedback set} We explore generating up to 16 edits per Feedback set. This means that given the prompt, original response, feedback, we generate up to 16 edited responses with temperature 0.7 and top p 0.9 and then pick the one with the highest reward from Llama-3.1-Nemotron-70B-Reward. Scaling from 2 to 16 edits increases Arena Hard from 88.4 (95\% CI: -1.7, 1.2) to 89.6 (95\% CI: -1.4, 1.3).

\paragraph{Scaling multiple axes simultaneously} Based on Fig. \ref{fig:front_page}, we chose to scale up responses per prompt and effective feedback per response, which seem to be more promising than edits per feedback set. Given compute constraints, we were only able to scale up to 8 responses with 16 effective feedback each. This means generating about 16 times as many tokens as one would with the default greedy generation, similar to the Best-of-N 16 responses setup. Our setup achieves an Arena Hard of 91.0 (95\% CI: -1.0, 1.2), meaning that it outperforms Best-of-N 16 at 88.5 (95\% CI: -1.2, 1.5). Finally, qualitative inspection suggests that Llama-3.1-Nemotron-70B-Reward model was not always optimal for selecting the best response. Therefore, after the conclusion of our initial experiments, we attempted to use an alternative recently-released Llama-3.3-Nemotron-70B-Select model \citep{nemotron70bselect} instead, which reaches Arena Hard of 92.7 (95\% CI: -1.2, 0.9). This surpasses both OpenAI o1-preview-2024-09-12 at 90.4 (95\% CI: -1.1, 1.3) \citep{arenahardauto} and DeepSeek R1 at 92.3 (95\% CI: unreported) \citep{deepseekai2025deepseekr1incentivizingreasoningcapability}.

\paragraph{Human Evaluation} To verify the effects of performance gains in automatic evaluation (Arena Hard), we also conduct a small-scale human-evaluation. We randomly identify 50 first-turn prompts from the General category that has not been previously used during HelpSteer3 annotations. Then, we ask 3 in-house annotators to independently annotate each prompt (8 annotators in total) on whether they prefer a response from Llama-3.1-Nemotron-70B-Instruct or the same model with an optimally-scaled Feedback-Edit ITS system (with randomly shuffled order to minimize position bias), following guidelines from \citet{wang2025helpsteer3preferenceopenhumanannotatedpreference}. We calculate the overall preference by averaging across three annotators and rounding to the closest preference rank. Out of 50 samples, 6 have at least one annotators indicating that both responses were invalid. Out of the remaining 44 samples, 13 responses from the ITS system were rated slightly better; 2 responses better and 1 response much better. 21 responses were rated about the same, 6 as slightly worse and 1 as worse. Overall, this shows that Feedback-Edit ITS is effective at 16 improvements vs. 7 degradations. Analysis of preference justifications reveal that increased response factuality, coherence and conciseness are the main contributors for such improvements.

\section{Distillation}

Inspired by \citet{deepseekai2025deepseekr1incentivizingreasoningcapability}, we investigate whether data generated from the Feedback-Edit system can be used to train models to generate better responses `zero-shot' (i.e. without using the scaling system at inference time). 

\paragraph{Data} 
We use all unique first-turn prompts from the Feedback Demonstration dataset (totaling 22, 644 with 21, 850 train and 794 val). We only use first-turn prompts due to practical constraints to minimize prefix lengths for generation throughput. In addition, we used vLLM \citep{kwon2023efficient}, a specialized inference framework in order to substantially increase generation throughput. We apply the optimal multi-axes scaling setup with Llama-3.1-Nemotron-70B-Instruct as the initial response model (Sec. \ref{sec:scaling}). This data is openly-released  \href{https://huggingface.co/datasets/nvidia/Llama-Nemotron-Post-Training-Dataset/viewer/SFT/chat?f[generator][value]=%27Llama-3.1-Nemotron-70B-Instruct,%20ITS%20models%27}{here}.

\paragraph{Model Training}

We use Llama 3.1 8B Instruct \citep{llama318b}, Llama 3.3 70B Instruct \citep{llama33} and Llama-3.1-Nemotron-70B-Instruct \citep{nemotron70binstruct} to initialize training. We perform SFT for 1 epoch (with 170 steps of global batch size of 128) and save checkpoints every 10 steps. The checkpoint with best validation performance is at steps 120, 150 and 160 respectively. Evaluation Metrics follows Section \ref{sec:evaluation} - Metrics. Further training details in Appendix \ref{appendix:training_details}.

\begin{table}[ht!]
\centering
\begin{adjustbox}{max width=\columnwidth, scale=1
}
\begin{tabular}{lcccc}
\toprule

& MT Bench  & Mean Response  & AlpacaEval  & Arena Hard \\
& (GPT-4-Turbo)  & Length (Chars.) & 2.0 LC (SE) & (95\% CI) \\
\midrule
\textbf{Llama-3.1-8B-Instruct}	& 7.43	& 1320.0 &	20.9 (1.25)	& 18.3 (-1.6, 1.6) \\
\midrule
+ Distillation	& 8.04	&3380.4	&41.5 (1.48)	& 55.5 (-2.3, 2.5) \\
\midrule
\textbf{Llama-3.3-70B-Instruct} & 8.29 & 1827.6 & 35.0 (1.45) & 62.4 (-2.5, 2.5) \\
\midrule
+ Distillation & 8.69 & 2837.5 & \textbf{61.6}  (1.21) & \textbf{88.8} (-1.7, 1.2) \\
\midrule
\textbf{Llama-3.1-Nemotron-70B-Instruct} & 8.98 & 2199.8 & 57.6 (1.65) & 85.0 (-1.5, 1.5) \\
\midrule
+ Distillation & \textbf{8.99} & 2508.2 & 61.3 (1.21) & 88.4 (-1.7, 1.6) \\

\bottomrule
\end{tabular}
\end{adjustbox}
\caption{Applying Distillation with various Instruct models. Higher is better for all metrics, except Length.
}
\label{tab:auto_eval_distillation}
\end{table}

\paragraph{Results}

Table \ref{tab:auto_eval_distillation} shows that Distillation can improve upon all initial models across MT-Bench, AlpacaEval and Arena Hard, although the improvements are generally lower compared to when the Feedback-Edit system is optimally applied at inference-time (see Table \ref{tab:auto_eval} and Fig. \ref{fig:front_page}). Notably, Llama-3.3-70B-Instruct has a substantial increase in AlpacaEval (35.0 to 61.6) and Arena Hard (62.4 to 88.8), indicating the effectiveness of the distillation process. Together, this suggests that the Feedback-Edit system can be useful for generating data to train models useful in applications that are latency-sensitive, while even greater benefits can be reaped when scaling at inference time. 

\section{Conclusion}

We show that dedicated Feedback and Edit models are useful for Inference-time Scaling in open-ended general-domain tasks, achieving 92.7 on Arena Hard and outperforming OpenAI o1-preview-2024-09-12 with 90.4 and DeepSeek R1 with 92.3, alongside improvements on MT Bench and AlpacaEval.

\section*{Ethical Considerations}

\paragraph{Societal Impact}

Beyond general alignment improvements discussed in the main paper, our Feedback-Edit Inference-Time Scaling approach brings along additional advantages that may enable more to develop and use advanced AI systems with higher customization, improved ease-of-deployment, lower latency and innate capacity for personalization.

First, the Feedback-Edit system offers fine-grained control over how much inference-time compute to use, by controlling the number of initial responses per prompt, feedback per response and edited responses per feedback set.

Second, Feedback and Edit models can be served in a disaggregated manner rather than on the same compute resource - which is necessary for monolithic models (e.g. DeepSeek R1 with over 600B parameters). 

Third, when performing scaling, sampling multiple candidates at each stage - initial responses, feedback and edited responses - can be done in parallel. This means that the overall system only has about 2 times the latency as greedy generation, since the initial response and edited response take similar time and feedback are much shorter and can be generated very quickly. On the other hand, thinking tokens from models such as DeepSeek R1 have to be generated sequentially. This leads to much higher latency upper-bounds, especially if a model thinks with many more tokens than it responds with, which is not uncommon.

Finally, using feedback that are human interpretable and selectable opens new research avenues for personalized alignment of models. Users can potentially choose model-generated feedback that is most in line with their own preferences. Such feedback can then be used to ground the Edit model to make changes based on individual preferences in a controllable fashion.

\paragraph{Licenses}
The use of Llama-3.1-8B-Instruct, Llama-3.1-Nemotron-70B-Instruct and Llama-3.1-Nemotron-70B-Reward is bound by the Llama 3.1 Community License Agreement while the use of Llama-3.3-70B-Instruct is bound by the Llama 3.3 Community License Agreement. We use all models in accordance with their license and intended use.

\paragraph{Human Annotations}
Annotators were contracted and managed by vendors Scale AI and Translated, which completed ethical review prior to the start of data collection. Consent for use of data for model training was obtained from all annotators through our vendors. Both Scale AI and Translated implemented several measures to prevent annotators from using LLMs as part of the annotation workflow.

Scale AI engages the Anker Methodology, GISC Impact Sourcing Standard, and UN Sustainable Development Goals to provide a fair and competitive pay. The specific pay is calculated based on many factors, including the specific project, the specialized skillset and expertise required, regional costs of living and then transparently listed on Scale AI platform. Scale AI also provides multiple channels for questions and support, including 24/7 support teams, community discussion channels with specially trained moderators, and a “speak up” hotline where contractors can report concerns anonymously. Worker concerns can be submitted to and are reviewed by the support team, and pay disputes are reviewed by support specialists trained in this area. 

Translated has active annotators on the platform earn at least their local minimum wage, with systems in place to ensure that annotators were paid in a timely manner for all the work they completed. Translated also has measures to avoid unreasonable levels of competition between annotators, as well as overwork. These measures include restricting sign-ups of new annotators as well as matching annotators with jobs on the platform, thereby reducing unpaid labour time spent on looking for jobs. In addition, Translated has active measures to protect annotators from work-related health and safety risks. Translated engages annotators with contracts that are written in clear language, accessible to annotators, consistent with annotators’ terms of engagement on the platform, and that contracts do not contain a clause requiring annotators to waive their right to legal recourse. Annotators can communicate with a human platform representative from Translated and there are officially documented and effective processes for annotators to appeal decisions such as bad reviews and ratings, or disciplinary actions. Translated also has a policy to mitigate the risk of discrimination against annotators by the platform or clients.

\section*{Limitations}

The current approach for sampling multiple feedback, filtering out feedback that find the response perfectly helpful and re-ranking feedback based on the number of words relating to constructive criticism is not compute-optimal. Techniques such as constrained decoding or involving other logit manipulation methods to incentivize decoding feedback with more constructive criticism may make such sampling more compute-efficient.

Data collection prompts sourced from ShareGPT \citep{sharegpt2023} and WildChat \citep{zhao2024wildchat1mchatgptinteraction} capture user queries collected before April 2023 and April 2024, respectively. The upper-bound of prompt complexity has since increased, as underlying systems have become more capable. Furthermore, ShareGPT and WildChat prompts requiring responses longer than 2000 words, or exceeding a 4,000 word input (across all user and assistant turns) were skipped altogether, as longer responses/prompts require more effort from annotators, which were outside of the scope that we had contractually agreed with our vendors. We also skipped prompts requiring access to knowledge after July 2024. Given this, results of this study may not fully demonstrate the types of complex or difficult queries that may benefit from Feedback and Edit Models . 

There is some risk that readers might interpret our Feedback-Edit system as the only useful method for Inference-Time Scaling. We believe this to be a misunderstanding of our contributions. Instead, we believe that different Inference-Time Scaling methods are beneficial for different settings and can complement each other in many scenarios. 

\bibliography{acl_latex}

\appendix

\newpage

\section{Prompt Preprocessing}\label{appendix:prompt_preprocessing}

\paragraph{Coding and Multilingual} We first identify the programming or natural language required for responding to each first-turn-user-prompt by prompting Nemotron 4 340B Instruct \citep{nvidia2024nemotron4340btechnicalreport} with prompt templates in Appendix \ref{appendix:prompt_templates}. Such identification is not always perfect and sometimes leads to confusion on the expected response - for instance responding with particular frameworks (e.g. Flask, Pandas or React) instead of the specific language (Python or Javascript). Prompts involving HTML, CSS or Javascript commonly require knowledge of all three languages and hence we combine into one `language' - Javascript/HTML/CSS. 

For natural languages, the LLM occasionally responses with the language name in the language itself (e.g. Español instead of Spanish). To solve such confusion, we manually map such occasions to the correct language name. Despite our best efforts, there are rare situations in which the language is misclassified and we ask our vendor to skip such tasks during annotations. Finally, we stratified sample prompts based on the language. Proportions are primarily determined based on the number of prompts available for each language in the ShareGPT dataset \citep{sharegpt2023}. This means that we allocate a higher proportion to commonly used languages such as Python and Chinese.

\paragraph{General and STEM} We first exclude prompts that involve languages other than English (directly or by mentioning the keyword translate), require code response, or are shorter than 10 characters since these messages tend to be some variant of Hi/Hello. We also exclude prompts that ask about the Chatbot's identity or contains keywords relating to OpenAI or ChatGPT, as these are commonly identity-related prompts that are not relevant for our purpose. As a note, we do not classify prompts into STEM and General prompts since STEM is a relatively broad term. Instead, our vendor manually classifies prompts and routes STEM prompts to suitable STEM specialist annotators subsequently. Following \citet{wang2024helpsteer2opensourcedatasettraining}, we classify the topic for each prompt using BERTopic \citep{grootendorst2022bertopic} into approximately 4,400 topics and predict the task complexity by prompting Nemotron 4 340B Instruct (see prompt template in Appendix \ref{appendix:prompt_templates}). Finally, we stratified sample prompts uniformly based on the topic and up-sample prompts that are more complex (either 4 or 5 out of 5 in terms of complexity).

\section{Prompt Templates}\label{appendix:prompt_templates}

Note: Newlines are removed in most appendices to due to compilation complication on LaTex.

\paragraph{Programming Language Identification}
\texttt{Please identify if expertise in any programming language is required to adequately address the following prompt. If there is more than one programming language needed, select the one that most relevant to the prompt. Provide only a one-word answer for the programming language if any is needed, or None otherwise. Here is the prompt: [prompt]}

\paragraph{Natural Language Identification}
\texttt{Please identify if knowledge of languages other than English is required to adequately address the following prompt. If there is more than one other language needed, select the one that most relevant to the prompt. Provide only a one-word answer for the language if any is needed, or None otherwise. Here is the prompt: [prompt]}

\paragraph{Prompt Complexity Prediction}

\texttt{Please evaluate the complexity of the following prompt based on the number of instructional intentions and the number of constraints. Provide a score between 1 and 5, where 1 represents very simple and straightforward, and 5 represents highly complex and intricate. Respond with this format only: [score]. Here is the prompt: [prompt]}

\paragraph{Feedback Application Identification}
\texttt{Change Summary: [change\_summary] Feedback: [feedback] Does the change summary address issues mentioned in the feedback? Answer only Yes or No}

\paragraph{Feedback Generation}
\texttt{[Entire Conversation] Evaluate the response to the previous prompt in terms of how helpful it is overall. Start the evaluation with the statement - The response is \{not / slightly / partially / mostly / perfectly\} helpful. Then provide a brief explanation of the evaluation in 2 to 10 sentences.}

\paragraph{Edit Generation} 
\texttt{[Entire Conversation] Edit the response to the previous prompt based on the following feedback: [feedback 1] [feedback 2] [feedback 3]} 

Please note that it uses up to 3 feedback where available. If no feedback is available, we use the filler \texttt{<None>} instead.

\paragraph{Edit Generation without Feedback} 
\texttt{[Entire Conversation] Edit the response to the previous prompt to improve it.}

\section{Bad Edit Details}\label{appendix:bad_edit_reasoning}

We found that this is due to annotators disagreeing with the provided feedback. When the research team inspected samples of disagreements, we found feedback annotators to be (much) more often correct compared to edit annotators - likely because edit annotators are dis-incentivized to identify issues that would result in major edits (i.e. more time/effort for editing) while feedback annotators are ambivalent to raising such issues.

\section{Further Dataset Statistics}\label{appendix:further_dataset_statistics}

\begin{table}[ht!]
\centering
\begin{adjustbox}{max width=\columnwidth, scale=1
}
\begin{tabular}{lccccc}
\toprule

\textbf{Programming Language} & Feedback & Edit & \textbf{Natural Language} & Feedback & Edit \\ 
Proportion & (\%)& (\%)& & (\%)& (\%) \\
\midrule

Python & 39.0 & 47.4 & Chinese & 29.6 & 32.4 \\
JavaScript/HTML/CSS & 22.6 & 19.6 & Korean & 11.2 & 10.4\\
SQL & 5.5 & 5.0  & French & 10.4 & 11.3 \\
Java & 5.2 & 5.4 & Spanish & 10.4 & 11.3\\
C\# & 5.1 & 4.6 & Russian & 6.8 & 5.4\\
C++ & 5.0 & 4.3 & Japanese & 6.7 & 6.0\\
Go & 3.7  & 2.4 & German & 5.6 & 5.1\\
C & 3.3 & 2.9 & Italian & 5.5 & 5.1\\
TypeScript & 3.2 & 2.4 & Portuguese & 5.3 & 4.8\\
PHP & 3.0 & 2.8 & Polish & 2.4 & 2.1\\
Bash & 1.2 & 0.7 & Indonesian & 2.1 & 1.9\\
Powershell & 1.1& 0.9 & Dutch & 2.0 & 2.1 \\
R & 1.1 & 0.8 & Vietnamese & 2.0 & 2.1 \\
Rust & 1.0 & 0.5 \\
\bottomrule
\end{tabular}
\end{adjustbox}
\caption{Proportion of Languages for Coding and Multilingual subsets for Feedback and Edit Demonstration Datasets.}
\label{tab:descriptive_feedback_languages}
\end{table}

Table \ref{tab:descriptive_feedback_languages} shows the proportion of responses from each language within the Coding and Multilingual subsets. There is a wide coverage of 14 programming and 13 natural languages, although both are over-represented by a few languages such as Python, Javascript/HTML/CSS and Chinese.

\begin{table}[]
\centering
\begin{adjustbox}{max width=\columnwidth, scale=1
}
\begin{tabular}{lcccccccccccc}
\toprule

\textbf{Country} & \multicolumn{2}{c|}{\textbf{General}}  & \multicolumn{2}{c|}{\textbf{STEM}} & \multicolumn{2}{c|}{\textbf{Coding}}  & \multicolumn{2}{c}{\textbf{Multilingual}} \\ 
Proportion (\%) & Feed. & Edit & Feed. & Edit & Feed. & Edit & Feed. & Edit\\
\midrule
AE & - & - & - & - & 0.2 & - & - & - \\
AR & 0.2 & 0.3 & 0.2 & 0.6 & 2.7 & 1.7 & 1.6 & 0.9 \\
AT & - & - & - & - & 0.0 & - & 1.3 & 0.7 \\
AU & 3.3 & 0.8 & 3.1 & 3.9 & 0.3 & 0.2 & 0.6 & 0.8 \\
BA & - & - & - & - & 0.8 & 0.2 & - & - \\
BD & - & - & - & - & 3.2 & 1.4 & - & - \\
BE & - & - & - & - & 0.0 & 0.1 & 0.5 & 0.5 \\
BJ & - & - & - & - & 0.0 & - & 2.1 & 0.3 \\
BO & - & - & - & - & 0.0 & 0.0 & 1.2 & 0.2 \\
BR & 0.4 & - & 0.4 & - & 0.8 & 1.2 & 3.5 & 3.0 \\
BZ & - & - & - & - & 0.1 & - & 0.2 & - \\
CA & 10.8 & 4.4 & 10.2 & 7.2 & 6.8 & 6.3 & 3.4 & 1.0 \\
CB & - & - & - & - & - & - & 1.0 & - \\
CH & - & - & - & - & 0.0 & - & - & - \\
CL & - & - & - & - & 0.4 & - & 1.6 & 2.2 \\
CM & - & - & - & - & 0.0 & - & 0.1 & 0.8 \\
CN & - & - & - & - & - & - & 12.9 & 14.0 \\
CO & 0.4 & - & 0.5 & - & 0.7 & 2.9 & 0.5 & - \\
CR & - & - & - & - & 0.1 & 0.1 & 0.7 & 0.2 \\
CY & - & - & - & - & - & - & 0.8 & 0.4 \\
CZ & - & - & - & - & 0.0 & 0.1 & 0.7 & - \\
DE & - & - & - & - & 0.1 & 0.0 & 5.0 & 5.3 \\
DK & - & - & - & - & 0.0 & - & 0.1 & - \\
EC & - & - & - & - & 0.1 & - & 0.3 & - \\
EE & - & - & - & - & 0.1 & - & - & - \\
EG & 0.0 & - & - & - & 15.4 & 18.1 & - & - \\
EL & - & - & - & - & - & - & 0.1 & - \\
ES & - & - & - & - & 0.4 & 0.1 & 3.4 & 7.3 \\
FI & - & - & - & - & 0.6 & 0.3 & 0.3 & 0.2 \\
FR & 0.0 & - & - & - & 0.3 & 0.1 & 5.0 & 6.9 \\
GB & 16.9 & 9.2 & 17.0 & 14.7 & 3.4 & 4.7 & 2.1 & 6.8 \\
GE & - & - & - & - & - & - & - & 0.4 \\
GR & - & - & - & - & 0.0 & 0.4 & 0.3 & 1.0 \\
HG & - & - & - & - & - & - & 0.9 & 0.4 \\
HK & - & - & - & - & - & - & 2.5 & 0.4 \\
HN & - & - & - & - & - & - & 0.2 & 0.9 \\
HU & - & - & - & - & 0.0 & 0.1 & 0.1 & 0.3 \\
ID & 0.1 & - & - & - & 0.9 & 0.4 & 1.9 & 1.9 \\
IE & 0.1 & 0.1 & 0.1 & 0.1 & 0.0 & 0.7 & 0.6 & 0.9 \\
IL & - & - & - & - & 0.4 & 0.7 & 0.3 & - \\
IN & 2.8 & 0.2 & 2.5 & - & 19.9 & 27.9 & 0.2 & - \\
IT & 0.0 & - & 0.0 & - & 0.0 & - & 5.0 & 7.1 \\
JO & 0.0 & - & - & - & 2.6 & 0.9 & - & - \\
JP & - & - & - & - & 0.0 & - & 4.4 & 4.7 \\
KE & - & - & - & - & - & - & 0.6 & 1.0 \\
KR & 0.0 & - & - & - & 0.0 & - & 7.1 & 8.4 \\
KZ & - & - & - & - & - & - & - & 0.4 \\
LT & - & - & - & - & - & - & 0.1 & - \\
LV & - & - & - & - & 0.0 & - & - & - \\
MA & - & - & - & - & 1.3 & 0.1 & - & - \\
MD & - & - & - & - & - & - & 0.5 & - \\
ME & - & - & - & - & - & 0.1 & - & - \\
MG & - & - & - & - & - & 0.1 & - & - \\
MX & 2.8 & 43.9 & 3.1 & 28.2 & 0.7 & 0.5 & 1.0 & 1.4 \\
MY & 0.3 & - & 0.5 & - & 0.0 & 0.2 & - & 0.6 \\
NG & - & - & - & - & - & - & 0.6 & - \\
NL & - & - & - & - & 0.0 & 0.1 & 0.5 & 1.2 \\
NZ & 0.8 & 0.1 & 0.9 & 0.3 & 0.5 & 0.0 & 0.1 & - \\
PE & - & - & - & - & 0.3 & 0.4 & - & 1.7 \\
PH & 3.2 & 0.5 & 2.1 & 0.7 & 1.2 & 1.2 & - & - \\
PL & - & - & - & - & 0.3 & 0.3 & 2.6 & 2.1 \\
PS & - & - & - & - & 4.7 & 0.8 & - & - \\
PT & - & - & - & - & 0.0 & 0.1 & 1.4 & 2.3 \\
PY & - & - & - & - & 0.0 & - & - & - \\
QA & - & - & - & - & 0.0 & - & - & - \\
RO & 0.0 & - & 0.1 & - & 0.8 & 0.2 & - & - \\
RS & - & - & - & - & 0.0 & - & 1.3 & - \\
SE & - & - & - & - & - & - & 0.2 & 0.6 \\
SG & - & - & - & - & 0.2 & - & 0.6 & 1.2 \\
TH & - & - & - & - & - & 0.0 & 0.3 & 0.4 \\
TJ & - & - & - & - & - & - & 0.1 & 0.2 \\
TR & 0.3 & 0.9 & 0.3 & 1.5 & 1.7 & 3.3 & 0.8 & 0.2 \\
TW & - & - & - & - & 0.0 & - & 6.0 & 3.2 \\
TZ & - & - & - & - & - & - & 0.4 & - \\
UA & - & - & - & - & - & - & 0.5 & 0.3 \\
UK & - & - & - & - & - & - & 2.1 & 0.2 \\
US & 57.3 & 39.6 & 58.9 & 42.7 & 25.3 & 24.0 & 5.6 & 3.1 \\
UY & - & - & - & - & 0.1 & 0.1 & 0.5 & - \\
VE & - & - & - & - & 0.3 & 0.0 & 0.0 & - \\
VN & - & - & - & - & 1.9 & 0.1 & 2.2 & 2.1 \\
ZA & 0.1 & - & 0.1 & - & 0.2 & 0.1 & - & - \\

\bottomrule
\end{tabular}
\end{adjustbox}
\caption{Proportion of Annotator Geographic Locations for each dataset, by alphabetic order of ISO-3166 Alpha-2 Code. 0.0 means exists but less than 0.05\%}
\label{tab:geographic}
\end{table}

\newpage

\section{Illustrative Example}\label{appendix:example}

\paragraph{Prompt:} Tourism marriage 

\paragraph{Initial Response:}  Tourism marriage, also known as ``tourism wedding'' or  ``destination wedding,'' refers to the practice of couples traveling to a different location, often a popular tourist destination, to get married. Here are some key aspects of tourism marriages: 1. **Destination**: Couples choose a destination that holds special meaning to them or is a place they've always wanted to visit. Popular destinations include beaches, mountains, historic sites, or even exotic locations abroad. 2. **Intimacy**: Tourism marriages often involve a smaller group of guests compared to traditional weddings. This can make the event more intimate and personal. 3. **Combination of Wedding and Honeymoon**: Many couples choose to have their honeymoon in the same location as their wedding, saving time and money. 4. **Unique Experiences**: Couples can incorporate local culture, traditions, or unique activities into their wedding, making it a memorable experience for everyone involved. 5. **Planning**: Planning a tourism marriage can be more complex than a local wedding. It often involves coordinating with vendors from afar, understanding local marriage laws, and managing travel logistics for guests. 6. **Cost**: Depending on the location and number of guests, a tourism marriage can sometimes be less expensive than a traditional wedding. However, it can also be more expensive, especially if the couple is covering travel costs for guests. 7. **Legal Considerations**: It's important for couples to understand the legal requirements for getting married in their chosen destination. Some countries have residency requirements or other regulations that could impact the wedding plans. 8. **Impact on Local Economy**: Tourism marriages can have a positive impact on the local economy, as they bring in visitors who will use local services and attractions. Here are a few popular destinations for tourism marriages: - Hawaii, USA - Bali, Indonesia - Tuscany, Italy - Cancun, Mexico - Santorini, Greece - The Caribbean Islands - Las Vegas, USA - Paris, France - Bora Bora, French Polynesia - Kyoto, Japan

\paragraph{Feedback:}
\begin{enumerate}
    \item The response is mostly helpful. Even though the prompt does not clearly specify the context, the model makes a good effort to return a sensible response by talking about what 'tourism marriages' or 'destination weddings' mean, 'key aspects of tourism marriages,' and 'some popular destinations for tourism marriages' as per recent trends.
    \item The response is mostly helpful. It defines tourism marriage, explains key aspects of tourism marriage, and names popular destinations. Some of the information the model chooses to include is generic and not necessary to write out - for example, the idea that traveling for a destination wedding can lead to  ``unique experiences'' as its own bullet point is not necessary.
    \item (Excluded to Edit Model) The response is perfectly helpful to the user. It explains what a tourism marriage is and gives other names that are used for it. The response gives several details that make up a tourism marriage and what they usually entail. Overall, the response is clear and understandable, giving the user a full explanation of the topic.
\end{enumerate}

\paragraph{Bad Edit (Modifications include changes not in scope of provided feedback):} Tourism marriage, also known as  ``tourism wedding'' or  ``destination wedding,'' refers to the practice of couples traveling to a different location, often a popular tourist destination, to get married. Here are some key aspects of tourism marriages: 1. **Destination**: Couples choose a destination that holds special meaning to them or is a place they've always wanted to visit. Popular destinations include beaches, mountains, historic sites, or even exotic locations abroad. 2. **Intimacy**: Tourism marriages often involve a smaller group of guests compared to traditional weddings. This can make the event more intimate and personal. 3. **Combination of Wedding and Honeymoon**: Many couples choose to have their honeymoon in the same location as their wedding, saving time and money. 4. **Planning**: Planning a tourism marriage can be more complex than a local wedding. It often involves coordinating with vendors from afar, understanding local marriage laws, and managing travel logistics for guests. 5. **Cost**: Depending on the location and number of guests, a tourism marriage can sometimes be less expensive than a traditional wedding. However, it can also be more expensive, especially if the couple is covering travel costs for guests. Local vendor pricing is also something to consider, as demand may be higher in popular destinations. 6. **Legal Considerations**: It's important for couples to understand the legal requirements for getting married in their chosen destination. Some countries have residency requirements or other complex regulations that could impact the wedding plans. Here are a few popular destinations for tourism marriages:  *   Hawaii, USA *   Bali, Indonesia *   Tuscany, Italy *   Cancun, Mexico *   Santorini, Greece *   The Caribbean Islands *   Las Vegas, USA *   Paris, France *   Bora Bora, French Polynesia *   Kyoto, Japan It is important to consider that certain countries may have different requirements when it comes to getting legally married. Many couples may choose to get legally married in their native country, and have a symbolic wedding abroad at their destination of choice, in order to avoid any difficult paperwork or wait for long periods of time for information to be processed.

\paragraph{Good Edit: (Modifications only include changes in scope of provided feedback)} Tourism marriage, also known as  ``tourism wedding'' or  ``destination wedding,'' refers to the practice of couples traveling to a different location, often a popular tourist destination, to get married. Here are some key aspects of tourism marriages: 1. **Destination**: Couples choose a destination that holds special meaning to them or is a place they've always wanted to visit. Popular destinations include beaches, mountains, historic sites, or even exotic locations abroad. 2. **Intimacy**: Tourism marriages often involve a smaller group of guests compared to traditional weddings. This can make the event more intimate and personal. 3. **Combination of Wedding and Honeymoon**: Many couples choose to have their honeymoon in the same location as their wedding, saving time and money. 4. **Planning**: Planning a tourism marriage can be more complex than a local wedding. It often involves coordinating with vendors from afar, understanding local marriage laws, and managing travel logistics for guests. 5. **Cost**: Depending on the location and number of guests, a tourism marriage can sometimes be less expensive than a traditional wedding. However, it can also be more expensive, especially if the couple is covering travel costs for guests. 6. **Legal Considerations**: It's important for couples to understand the legal requirements for getting married in their chosen destination. Some countries have residency requirements or other regulations that could impact the wedding plans. 7. **Impact on Local Economy**: Tourism marriages can have a positive impact on the local economy, as they bring in visitors who will use local services and attractions. Here are a few popular destinations for tourism marriages:  * Hawaii, USA * Bali, Indonesia * Tuscany, Italy * Cancun, Mexico * Santorini, Greece * The Caribbean Islands * Las Vegas, USA * Paris, France * Bora Bora, French Polynesia * Kyoto, Japan

\section{Aligned Model Evaluation Details}\label{appendix:aligned_model_evaluation}

\paragraph{MT Bench} We follow \citet{meng2024simpo, tenyxchat2024, wang2024helpsteer2preferencecomplementingratingspreferences} to use MT Bench \citep{zheng2023judging}, with GPT-4-Turbo (specifically GPT-4-0125-Preview) as the judge. MT Bench consists of 80 multi-turn questions, each consisting of an initial question and a follow-up question, for a total of 160 prompts. Questions are collated from 8 categories: Reasoning, Math, Coding, STEM, Writing, Roleplay, Extraction and Humanities/Social Science. Responses are first generated greedily with up to 1024 tokens (default value for MT Bench). The responses to these prompts are evaluated by GPT-4-0125-Preview to give a score between 1 and 10, and the mean across all prompts is reported. Prompts in Coding, Math and Reasoning categories are evaluated with a reference correct answer, which were generated by the judge model and then manually verified to be correct. Higher MT Bench score indicates better instruction following ability.

\textbf{AlpacaEval 2.0 Length Controlled} We follow \citet{dong2024rlhf, meng2024simpo, wang2024helpsteer2preferencecomplementingratingspreferences} to use AlpacaEval 2.0 Length Controlled \citep{dubois2024length}. AlpacaEval 2.0 contains 805 first-turn instructions (relating to singular-requirement, straightforward tasks such as recommendations, question answering, and open-ended writing). An answer to each prompt is greedily generated up to 2048 tokens by the evaluated model as well as a baseline model (GPT-4-1106-turbo), which are then sent to GPT-4-1106-turbo evaluator that outputs the probability of preferring the generations of the evaluated model. Finally, the authors introduced a length correction factor to mitigate the bias for the evaluator towards preferring longer generations.

\textbf{Arena Hard} We follow \citet{dong2024rlhf, meng2024simpo, wang2024helpsteer2preferencecomplementingratingspreferences} to use Arena Hard \citep{arenahard2024}. Arena Hard contains 500 first-turn instructions obtained from challenging user queries on Chatbot Arena \citep{chiang2024chatbot}. Prompts are classified using an LLM (Llama-3-70B-Instruct) to determine if they are complex, specific, real-world-application-related or require domain knowledge, problem-solving, creativity, technical accuracy. Prompts that meet at least 6 out of 7 criteria are labelled as challenging. Therefore,  a huge proportion of prompts (>50\%) are related to solving coding problems. There is also a small fraction of samples involving multlingual capabilities. An answer to each prompt is greedily generated up to 2048 tokens by the evaluated model. Model responses are then compared with responses from GPT-4-0314 using GPT-4-1106-preview judge to calculate a win-rate over GPT-4-0314.

\section{Training Details}\label{appendix:training_details} 
All experiments were conducted using the Nemo-Aligner framework \citep{shen2024nemoaligner}, specifically with the container (nvcr.io/nvidia/nemo:24.05.llama3.1).
We train models with a max sequence length of 4096 and discard any data longer than this, which only consists of less than 3\% of each dataset. In all experiments, we use  AdamW optimizer \citep{loshchilov2017decoupled} with 10 warmup steps.
Across SFT, RM and Distillation experiments, we conduct search over constant learning rates over $\{1, 2, 3\}e-6$. For both Feedback, Edit SFT and 8B Distillation, we found $1e-6$ to work the best while for Edit RM and 70B Distillation, $3e-6$ works the best. For Edit RL, we only tried constant LR of $5e-7$ and KL of $0.01$ following \citet{wang2024helpsteer2preferencecomplementingratingspreferences} since RLOO is not sensitive to hyper-parameters. In addition, we rollout on 4 responses per prompt with up to 2048 generated tokens with temperature 1 and top p 0.9 following \citet{wang2024helpsteer2preferencecomplementingratingspreferences}. vLLM \citep{kwon2023efficient}
Inference uses the 0.6.6.post1 container.

\section{Compute requirements}\label{appendix:compute_requirements}

\begin{table}[ht!]
\centering
\begin{adjustbox}{max width=\columnwidth, scale=1
}
\begin{tabular}{lccccc}
\toprule
\textit{Stage} &  \multicolumn{3}{c}{Compute (H100-eqv. node-hours)}  \\
\midrule
& SFT & RM & RL &  \\
\midrule
Feedback & 80 & - & -\\
Edit & 16 & 5 & 48\\
Distillation & 16 & - &- \\
\bottomrule
\end{tabular}
\end{adjustbox}
\caption{Compute required for training 70B models in H100 node-hours. Experiments are run on nodes of 8 H100-80GB SXM GPUs on internal clusters.}
\label{tab:compute}
\end{table}

\section{Example Feedback by prompting Llama-3.1-Nemotron-70B-Instruct}\label{appendix:llama31nemotron_feedback}

\textbf{Note: Below is one single feedback}

The response is partially helpful.

This evaluation is based on the response's strengths and weaknesses in addressing the prompt for a literature review on AI and Patient Care Optimization. On the positive side, the response:

Provides a clear and concise structure for a literature review, including an introduction, themed sections, and a conclusion.

Covers a range of relevant topics within the field, such as diagnostic accuracy, personalized medicine, workflow optimization, patient engagement, and ethical considerations.

Includes a formatted reference list, albeit with fictional citations due to the nature of the prompt.

However, the response falls short in several key areas, limiting its overall helpfulness:

Lack of Real-World Citations: The use of fictional citations (e.g., "Char, D. S., et al. (2018)") hinders the response's practical utility, as readers cannot pursue the referenced studies.

Superficial Depth: The discussion of each topic is brief and lacks the in-depth analysis typically expected in a literature review, failing to provide nuanced insights or critically evaluate the studies.

Absence of Methodological Details: The response does not outline the methodology used for the literature search, selection criteria, or how the themes were identified, which are crucial components of a comprehensive literature review.

No Clear Implications or Future Research Directions: Beyond a general call for addressing ethical challenges, the response does not clearly articulate implications for practice, policy, or future research avenues, which would enhance its helpfulness for readers seeking to apply the knowledge.

\newpage
\section{Example Feedback by Feedback Model}\label{appendix:example_feedback}

\textbf{Below are three feedback to the same response. Constructive criticism highlighted in bold.}

\begin{enumerate}
    \item The response is mostly helpful. The response provides a thorough literature review of AI in patient care optimization, effectively synthesizing relevant studies to highlight key themes. The citations are correctly formatted, and the content is well-organized. \textbf{Minor redundancies exist, but} the overall quality and completeness make it a valuable resource for understanding the current state of AI in healthcare.
    \item The response is partially helpful. The response provides a literature review of AI and patient care optimization. The response uses the proper citation format, and provides a detailed review of the topic. \textbf{However, the response is lengthy and could be more concise. The response could also have more information on the challenges of AI in patient care.}
    \item The response is mostly helpful. The response provides a comprehensive literature review on AI and patient care optimization. It covers various aspects such as predictive analytics, personalized medicine, and challenges in implementation. The response is well-structured and easy to follow. \textbf{However, some of the references are not directly related to patient care optimization.}
\end{enumerate}
\newpage
\section{Additional Comparisons}\label{appendix:additional_baselines}

\begin{table}[ht!]
\centering
\begin{adjustbox}{max width=\columnwidth, scale=1
}
\begin{tabular}{lcccc}
\toprule

& MT Bench  & Mean Response  & AlpacaEval  & Arena Hard \\
& (GPT-4-Turbo)  & Length (Chars.) & 2.0 LC (SE) & (95\% CI) \\

\midrule
\textbf{Llama-3.1-Nemotron-70B-Instruct} & 8.98 & 2199.8 & 57.6 (1.65) & 85.0 (-1.5, 1.5) \\
\midrule
+ Thinking Step by Step &9.00 &2702.9&49.9 (1.50)&81.2 (-1.8, 1.6) \\
+ CritiqueLLM + Self-Edit	& 7.96 &3129.0	&66.0 (1.25)	& 82.2 (-1.6, 2.2)\\
+ Feedback + Self-Edit & 8.94 & 2320.6 & \textbf{66.2} (1.40) & 85.4 (-1.6, 1.4)\\
+ Feedback + Edit & \textbf{9.16} & 2614.4 & 62.8 (1.30) & \textbf{87.0} (-1.5, 1.7) \\
\midrule
\textbf{Gemma-2-27B-IT}	& 8.42 &	1248.0	& 55.9 (1.50)	& 47.5 (-2.5, 2.7) \\ 
\midrule
+ Feedback + Edit	& 8.54	& 1845.6	& 54.1 (1.50)	& 76.2 (-2.3, 2.3) \\
\midrule
\midrule
\textbf{Effect of Feedback/Edit Model Size} \\
\midrule
\midrule
\textbf{Llama-3.1-8B-Instruct} & 7.43 & 1320.0 & 20.9 (1.25)	& 18.3 (-1.6, 1.6) \\
+ 8B Feedback + 8B Edit w/o RL	& 7.34	& 1492.6 & 16.9 (1.15)	& 15.8 (-1.5, 1.5) \\ 
+ (70B) Feedback + (70B) Edit w/o RL & 	8.16 & 	1671.6	& 19.8 (1.24) & 34.0 (-2.4, 2.5) \\

\textbf{Llama-3.1-Nemotron-70B-Instruct} &	8.98	& 2199.8	& 57.6 (1.65) &	85.0 (-1.5, 1.5) \\
+ 8B Feedback + 8B Edit w/o RL	& 8.76 &	2261.4	& 63.2 (1.35) &	83.9 (-1.7, 1.7) \\
+ (70B) Feedback + (70B) Edit w/o RL	& 9.12 &	2284.4	& 64.4 (1.35) & 	86.4 (-1.5, 1.5) \\

\bottomrule
\end{tabular}
\end{adjustbox}
\caption{Additional Comparisons for Applying Feedback and Edit models with various Instruct models. Higher is better for all metrics, except Length.
}
\label{tab:additional_eval}
\end{table}

We conduct further comparisons at the recommendation of anonymous reviewers, whom we are grateful towards.

\paragraph{Let's Think Step by Step} As shown in Table \ref{tab:additional_eval}, prompting with Chain-of-Thought \citep{wei2023chainofthoughtpromptingelicitsreasoning} (e.g. “Let’s think step by step”) maintains the MT Bench performance but substantially degrades performance on AlpacaEval 2 LC and Arena Hard, relative to the base Llama-3.1-Nemotron-70B-Instruct model.

\paragraph{CritiqueLLM}
We also generate feedback using CritiqueLLM \citep{ke-etal-2024-critiquellm} (based on their released checkpoint and inference code \href{https://github.com/thu-coai/CritiqueLLM/tree/main}{here}) before generating edits using Self-Edit by the Llama-3.1-Nemotron-70B-Instruct model. As our Edit model is not directly compatible with CritiqueLLM feedback, we evaluate  with Self-Edit only. We also notice that the feedback generated by CritiqueLLM is always in Chinese and simply prompting Llama-3.1-Nemotron-70B-Instruct to Self-Edit with the original prompt cause the responses to be also in Chinese, despite the benchmarks expecting responses in English. To mitigate potential issues due to this language mismatch, we add a further instruction to “Please write the edited response in English.” after the Self-Edit instruction and feedback. Table \ref{tab:additional_eval} shows that while there is improvement in AlpacaEval from 57.6 to 66.0 by using CritiqueLLM, MT Bench drop drastically (8.98 to 7.96) while Arena Hard decreases from 85.0 to 82.2. The drop in MT Bench and Arena Hard relative to using Llama-3.1-Nemotron-70B-Instruct alone, suggests that the feedback provided by CritiqueLLM coupled with Self-Edit does not achieve the same gains as our dedicated Feedback and Edit models (which were able to make substantial improvements on all three benchmarks). Furthermore, when using our Feedback model with Self-Edit, we still get higher scores in each benchmark than CritiqueLLM with Self-Edit.

\paragraph{Gemma-2-27B-IT} To show the effects of the Feedback-Edit models outside of Llama based models, we also utilize the Feedback and Edit models on Gemma 2 27B Instruct \citep{gemmateam2024gemma2improvingopen} (used in accordance with Gemma Terms of Use). Table \ref{tab:additional_eval} shows that when augmented with a Feedback and Edit system, the model substantially improves in terms of MT bench and Arena Hard while maintaining its performance in AlpacaEval (within Standard Error of each other).

\paragraph{Effect of Feedback/Edit Model Size} 

To better understand the effect of model size on the performance of Feedback and Edit Models (70B in main experiments), we also train 8B-sized Feedback and Edit models (initialized with Llama-3.1-8B-Instruct). We only train these without RL as RL requires substantially more resources and pre-RL results do not indicate promising results for 8B models.
Table \ref{tab:additional_eval} shows that 8B Feedback and Edit models are generally unable to improve both 8B and 70B starting models and many metrics indicate substantial degradation relative to the starting model alone (e.g. 8B: AlpacaEval; 70B: MT Bench). The only observed improvement is AlpacaEval 2 LC for the 70B starting model but the extent of improvement is less than the (+70B Feedback + 70B Edit w/o RL) baseline. On the other hand, 70B Feedback and Edit models are able to make improvements across all metrics across 8B and 70B starting models, except for AlpacaEval 2 LC for 8B starting model, which it maintains at a similar level. This suggests that a model with a larger number of parameters (70B) is useful to provide high quality feedback and edits while a similarly trained 8B model is less useful. For 8B-sized models, distillation with 70B Feedback and Edit models can be a more promising approach, as indicated in Table \ref{tab:auto_eval_distillation}.

\end{document}